\documentclass[conference]{IEEEtran}
\IEEEoverridecommandlockouts
\usepackage{cite}
\usepackage{amsmath,amssymb,amsfonts}
\usepackage{algorithmic}
\usepackage{graphicx}
\usepackage{textcomp}
\usepackage{xcolor}

\usepackage{booktabs}
\usepackage[ruled,vlined]{algorithm2e}
\usepackage{multirow}
\usepackage{soul}
\usepackage{bbm}
\usepackage{amsthm}
\usepackage{subcaption}

\newtheorem{theorem}{Theorem}

\newtheorem{definition}[theorem]{Definition}

\def\BibTeX{{\rm B\kern-.05em{\sc i\kern-.025em b}\kern-.08em
    T\kern-.1667em\lower.7ex\hbox{E}\kern-.125emX}}
\begin{document}

\title{Multi-Grained Temporal-Spatial Graph Learning for Stable Traffic Flow Forecasting}

\author{\IEEEauthorblockN{Zhenan Lin\IEEEauthorrefmark{1}, Yuni Lai\IEEEauthorrefmark{2}\thanks{Mr. Zhenan Lin and Dr. Yuni Lai provide equal contributions to this project.}, Wai Lun Lo\IEEEauthorrefmark{1}, Richard Tai-Chiu Hsung\IEEEauthorrefmark{1}, \\Harris Sik-Ho Tsang\IEEEauthorrefmark{1}, Xiaoyu Xue\IEEEauthorrefmark{2}, Kai Zhou\IEEEauthorrefmark{2}, Yulin Zhu\IEEEauthorrefmark{1}\thanks{Dr. Yulin Zhu is the corresponding author.}}
\IEEEauthorblockA{\IEEEauthorrefmark{1}\textit{Department of Computer Science}, \textit{Hong Kong Chu Hai College}, Hong Kong, China\\ 
246360250@student.chuhai.edu.hk, wllo@chuhai.edu.hk, richardhsung@chuhai.edu.hk, \\harristsang@chuhai.edu.hk, ylzhu@chuhai.edu.hk}
\IEEEauthorblockA{\IEEEauthorrefmark{2}\textit{Department of Computing}, \textit{The Hong Kong Polytechnic University}, Hong Kong, China\\
cs-yuni.lai@polyu.edu.hk, xiaoyu.xue@connect.polyu.hk, kaizhou@polyu.edu.hk}
}

\maketitle


\begin{abstract}
    Time-evolving traffic flow forecasting are playing a vital role in intelligent transportation systems and smart cities. However, the dynamic traffic flow forecasting is a highly nonlinear problem with complex temporal-spatial dependencies. Although the existing methods has provided great contributions to mine the temporal-spatial patterns in the complex traffic networks, they fail to encode the globally temporal-spatial patterns and are prone to overfit on the pre-defined geographical correlations, and thus hinder the model's robustness on the complex traffic environment. To tackle this issue, in this work, we proposed a multi-grained temporal-spatial graph learning framework to adaptively augment the globally temporal-spatial patterns obtained from a crafted graph transformer encoder with the local patterns from the graph convolution by a crafted gated fusion unit with residual connection techniques. Under these circumstances, our proposed model can mine the hidden global temporal-spatial relations between each monitor stations and balance the relative importance of local and global temporal-spatial patterns. Experiment results demonstrate the strong representation capability of our proposed method and our model consistently outperforms other strong baselines on various real-world traffic networks.     
    
\end{abstract}


\section{Introduction}
\label{Sec-intro}

The rapid urbanization and the increasing number of vehicles have intensified the challenges of urban traffic management, resulting in significant economic, environmental, and social implications~\cite{ahmed1979analysis,van2005accurate,williams2001multivariate}. Accurate traffic flow forecasting is a critical component of Intelligent Transportation Systems (ITS), enabling applications such as dynamic traffic signal control, route optimization, congestion management, and preemptive resource allocation~\cite{li2024survey,jiang2022graph,jiang2023graph}. However, traffic flow prediction is inherently complex due to the spatio-temporal dependencies and dynamic nature of traffic patterns, influenced by factors such as weather, special events, accidents, and infrastructure changes. Addressing these challenges requires advanced modeling techniques capable of capturing the intricate interplay between spatial and temporal dependencies in traffic networks.

In recent years, Deep Learning (DL) has emerged as a transformative approach to traffic forecasting, offering significant improvements over traditional statistical and machine learning methods~\cite{williams2003modeling,LSVR,LSVR2,adaboost,gradientboosting,zhao2016high,pan2013short,sun2006bayesian}. Classical methods such as Autoregressive Integrated Moving Average (ARIMA)~\cite{williams2003modeling} and Kalman filters suffer from limitations in handling non-linearity and high-dimensional data. Machine learning approaches, including Support Vector Regression (SVR)~\cite{LSVR,LSVR2} and Random Forests~\cite{adaboost}, improve upon these limitations but often require extensive manual feature engineering and fail to effectively model the spatial dependencies inherent in traffic networks. Deep learning models~\cite{lv2014traffic,LSTM,fu2016using,yu2017spatiotemporal,shi2015convolutional,zhang2017deep,ke2017short,wu2016short,DCRNN,guo2019attention,wu2019graph,vaswani2017attention,zheng2020gman,xu2020spatial,hu2023multi,cai2020traffic,song2020spatial,guo2020short,park2020st,li2017diffusion}, particularly those leveraging Recurrent Neural Networks (RNNs)~\cite{li2017diffusion,yu2017spatiotemporal} and Convolutional Neural Networks (CNNs)~\cite{pmlr-v48-niepert16,Li_Cui_Zheng_Xu_Yang_2018,10.5555/3157382.3157527,shi2015convolutional}, have demonstrated the ability to automatically learn patterns from complex, high-dimensional data. Furthermore, Graph Neural Networks (GNNs)~\cite{guo2019attention,song2020spatial,STGCN} and Transformer-based architectures~\cite{xu2020spatial,hu2023multi,cai2020traffic} have shown great promise in capturing both local and global dependencies in graph-structured data, such as traffic networks.

\begin{figure*}[h]
	\centering
    \includegraphics[width=0.85\textwidth,height=7cm]{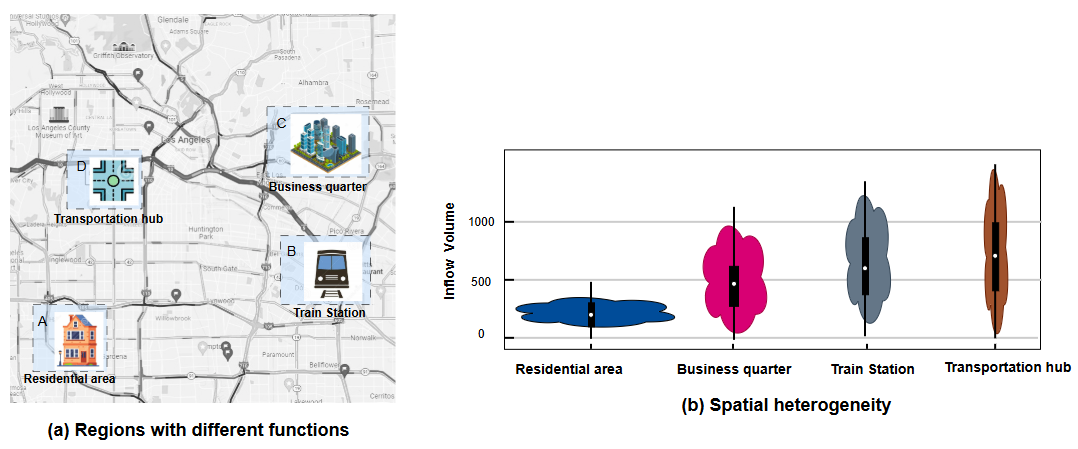}
	\caption{Spatial heterogeneity in traffic graph dataset.}
    \label{fig-spatial-heter}
\end{figure*}

Despite these advancements, several critical challenges remain unaddressed in traffic forecasting using deep learning. Many existing methods~\cite{li2024survey, jiang2022graph, lv2014traffic} focus on either local spatial-temporal dependencies or global relationships, neglecting the need to balance and adaptively integrate these patterns. This results in suboptimal representation of traffic data and limits the predictive accuracy. Furthermore, pre-defined geographical correlations used in some models~\cite{STGCN, shi2015convolutional, zhang2017deep, wu2016short} can lead to overfitting, reducing their generalization in real-world dynamic traffic environments. These gaps highlight the need for innovative hybrid architectures that can adaptively combine local and global temporal-spatial patterns and generalize across diverse traffic scenarios.

This study proposes a novel multi-grained graph learning framework that combines the temporal-spatial graph convolutional blocks with a graph enhanced transformer module. The hybrid architecture leverages the complementary strengths of both modules: the temporal-spatial message passing mechanism of the graph convolutional layer excels at capturing localized temporal-spatial dependencies, while the graph enhanced transformer module effectively models long-range global relationships, which are not easily perceived directly in the geographical information. For example, in Fig.~\ref{fig-spatial-heter}, there are plenty of unconnected traffic nodes share the similar urban functions such as ``transportation hub", ``business quarter", ``residential area", ``train station", etc. However, the localized message-passing mechanism of the GCN encoder will easily omit the urban function similarity (semantic information in the node attributes) of the long-range traffic node pair and thus cause detrimental effects to the temporal-spatial graph representation learning. To further enhance the model's performance, we novelly introduce the adaptive gated units, which dynamically adjust the relative contributions of local temporal-spatial patterns captured by the temporal-spatial graph convolutional blocks and global temporal-spatial patterns extracted by the graph-enhanced transformer module. This adaptive mechanism allows the model to balance the two types of dependencies, enhancing its expressive power and enabling it to effectively capture both micro-level traffic trends and macro-level global relationships. Furthermore, the model incorporates a crafted residual connection technique to facilitate the flow of information across layers, thereby improving the stability and efficiency of training.

The proposed model is evaluated on real-world dynamic traffic datasets, including PeMSD4 and PeMSD8~\cite{PeMSD4}, which represent diverse traffic conditions and network structures. Experimental results demonstrate that the model consistently outperforms existing state-of-the-art methods, highlighting its ability to effectively capture both local and global spatio-temporal dependencies in complex traffic environments.

This study makes several key contributions. First, it introduces a novel hybrid architecture that combines temporal-spatial graph convolutional blocks and the graph enhanced transformer module with an adaptive fusion mechanism, allowing for the dynamic integration of local and global temporal-spatial patterns. Second, the proposed model incorporates meticulously designed components, including gated residual connections in the temporal-spatial graph convolutional blocks and spatial-aware self-attention in the graph enhanced transformer layer, to effectively capture comprehensive spatio-temporal dependencies. Lastly, extensive experimental results demonstrate the model’s superior predictive performance across a variety of traffic networks, underscoring its practical applicability to real-world intelligent transportation systems. By addressing critical gaps in the literature and advancing the state-of-the-art in traffic flow forecasting, this study contributes to the development of more reliable and accurate ITS, ultimately paving the way for smarter, more sustainable urban transportation systems.

The remainder of this paper is organized as follows: Section~II reviews related work, highlighting the evolution of traffic forecasting methods and recent advancements in deep learning. Section~III provides the preliminary by defining the traffic flow graph and the graph-based traffic flow prediction problem. Section~IV details the proposed methodology, including the hybrid architecture, adaptive gated units, and adversarial training strategy. Section~V describes the experimental setup and presents the results, including a comparison with strong baselines and ablation studies. Finally, Section~VI concludes the paper by summarizing the findings and discussing future research directions.

\section{Related Works}
\subsection{Traffic Prediction}
In recent years, many efforts have been contributed to promote the traffic flow predictive models. The traditional statistical models involve in traffic flow prediction are HA, ARIMA~\cite{williams2003modeling,williams2001multivariate}, LSVR~\cite{LSVR,LSVR2}, etc. HA represents the historical average method and predict the traffic flow information based on the average of the proximal traffic flow features. ARIMA is a vanilla statistical time series model that predicts traffic flows based on their inherent periodic property. LSVR is SVR with a linear kernel and maps the traffic flows into a linearized Hilbert space. However, all the above-mentioned methods highly rely on strong assumptions and cannot capture the complex nonlinear temporal-spatial patterns. Later, some deep-learning-based methods have been introduced to tackle the traffic flow prediction problem. For example, LSTM~\cite{LSTM} utilizes the special gate mechanism and memory cell units to capture the complex nonlinear temporal relationship between the station pair. Yao et al.~\cite{DMVST} proposed a deep multiview spatial-temporal network that integrates convolutional neural networks and LSTM to jointly model both spatial and temporal dependencies to predict large-scale taxi demand. However, the limitation of those deep-learning-based methods is their inadequate capability on capturing the geographical information of the traffic data compared to the graph-based models.   
Later, DCRNN~\cite{DCRNN} regards the traffic flow prediction as a diffusion process on a directed graph and captures the spatial dependency using bidirectional random walks on the graph, and the temporal dependency using the encoder-decoder architecture with scheduled sampling.  

\subsection{Temporal-Spatial Graph Convolution}
The conventional convolutional neural networks~\cite{yu2017spatiotemporal,shi2015convolutional} mine the geographical relationship in the traffic data by partitioning the traffic map into a series of grids, which can only provide coarse spatial correlations and lead to suboptimal performances. However, in recent years, lots of researchers determine to curve the inherent geographical connections for each transport node by encoding such information into a topology space and adopt the message passing mechanism of the graph convolutional layer to mine the fine-grained geographical relations in the traffic data. Two mainstreams of graph convolution methods~\cite{STGCN,Li_Cui_Zheng_Xu_Yang_2018,pmlr-v48-niepert16,guo2019attention,song2020spatial,10.5555/3157382.3157527} are the spatial methods and the spectral methods. The spatial methods utilize the message-passing mechanism to propagate the traffic information between the target node with its neighbors. For example, Niepert and et al.~\cite{pmlr-v48-niepert16} adopted a linear method to select the neighborhood of every target node to supervise the information propagation on the topology space. Li and et al.~\cite{Li_Cui_Zheng_Xu_Yang_2018} introduced a spatial-temporal graph convolution approach to capture the spatial information of the human skeleton-based actions. As for the spectral methods, researchers analyze the relational data via the spectrum space and utilize the graph Laplacian to measure the closeness between each node. For example, Defferrard and et al.~\cite{10.5555/3157382.3157527} utilized the Chebyshev polynomial approximation to design the fast localized convolutional filters on graph data. Yu and et al.~\cite{STGCN} introduced the spatial-temporal convolution blocks to simultaneously capture the inherent temporal spatial connections in the dynamic traffic graph data. However, the above-mentioned GNN-based methods are prone to overfit on the localized temporal-spatial relations on the traffic data and thus converge to suboptimal results.

\section{Preliminary}
In this section, we will introduce the definition of the vital notations and the graph-based traffic flow prediction task. 

\begin{definition}
(Traffic Flow Graphs): A traffic flow graph is defined as $\mathcal{G}_t=(\mathcal{V}_t, \mathcal{E}_t, \mathbf{X}_{t})$, where $\mathcal{V}_t=\{v^i_t\}_{i=1}^{N}$ is the node set containing a series of nodes representing the observations from $N$ monitor stations in the traffic flow graphs; $\mathcal{E}_t=\{e^i_t\}_{i=1}^{E}$ is a set of links representing the connection between each monitor station. $\mathbf{X}_t$ indicates the traffic flow related features for each monitor station at $t$-th time step. 
\end{definition}
In general, the traffic flow information (such as total volume, average speed, etc.) is encoded in the attribute matrices $\{\mathbf{X}_t\}_{t=1}^{T}$.
Without loss of generality, in this paper, we use the adjacency matrix $\mathbf{A}_t=\{\mathbf{A}_{t}^{ij}\}_{i,j=1}^{N}$ to represent the topology information of the traffic flow graph in the $t$-th time step. During the graph construction phase, we usually utilize the Gaussian-kernel-based similarity scores to represent the connection strength between each monitor station:
\begin{equation}
     \mathbf{A}_{t}^{ij}=
\begin{cases} 
\exp\{-\frac{(d_t^{ij})^2}{\sigma^2}\},  & \mbox{if }i\neq j\mbox{ and } \exp\{-\frac{(d_t^{ij})^2}{\sigma^2}\}\geq\epsilon, \\
0, & \mbox{otherwise}, \label{eqn-data-preprocessing}
\end{cases}
\end{equation}
where $d_t^{ij}$ is the real distance on the road map between the monitor station $i$ and $j$ at $t$-th time step; $\sigma^2$ is the hyperparameter that controls the shape of the Gaussian distribution, and thus indirectly determine the scale of the similarity metric. For example, a small $\sigma^2$ indicates that only sufficient high value of $\mathbf{A}_{t}^{ij}$ can be regarded as high similarity. On the other hand, $\epsilon$ is another hyperparameter that controls the sparsity of the adjacency matrix. A large $\epsilon$ leads to a sparse traffic flow graph.

\begin{definition}
    (Graph-based Traffic Flow Prediction) Given the historical traffic flow graphs within a window size with length equal to $\tau+1$:
    $\mathcal{G}_{\{T-\tau:T\}}=\{\mathcal{G}_{T-\tau},\mathcal{G}_{T-\tau+1},...,\mathcal{G}_T\}$, 
    we aim to learn an (sub-)optimal forecasting function $f^*$ to precisely fit the traffic flow information at time step $T+1$, $T+2$,..., $T+\tau$, 
    i.e., 
    \begin{equation}
        \hat{\mathbf{X}}_{T+1},...,\hat{\mathbf{X}}_{T+\tau}=f^*(\mathcal{G}_{T-M+1},...,\mathcal{G}_{T}),
    \end{equation}  
    where $\hat{\mathbf{X}}_T$ 
    is the prediction on the traffic flow information at $T$-th time step, $M$ is the length of the window size used for collecting the historical data.  
\end{definition}

\begin{table}[h]
	\centering
	\caption{Summary of Notations.}
	\label{tab-notations}
	\resizebox{1.\columnwidth}{!}{%
		\begin{tabular}{c|c}
			\toprule[1.pt]
			Symbol & Description \\
			\hline
			$\mathcal{G}_t$ & Traffic graph data at $t$-th time step\\
            $\mathcal{V}_t$ & Node set for the traffic graph data at $t$-th time step\\
            $\mathcal{E}_t$ & Link set for the traffic graph data at $t$-th time step\\
            $\mathbf{X}_t$ & Nodal attributed matrix at $t$-th time step\\
            $\mathbf{A}_t$ & Adjacency matrix at $t$-th time step\\
            $\hat{\mathbf{X}}_t$ & Predictions on the traffic flow at $t$-th time step\\
            $\mathbf{W}^{(l)}$ & Weight matrix of the $l$-th graph convolutional layer\\
            $C_T(\cdot)$ & Gated linear units operation\\
            $C_S(\cdot)$ & Graph convolution operations\\
            $\sigma(\cdot)$ & Sigmoid activation function\\
            $P_{t-\tau:t}$ & Output features from the 1-D causal convolutions\\
            $Q_{t-\tau:t}$ & Output features from the 1-D causal convolutions\\
            $\mathbf{L}_t$ & Node embeddings for locally temporal-spatial aggregation\\
            $\mathbf{D}_t$ & Degree matrix for locally temporal-spatial aggregation\\
            $z^{-}_{deg^{-}(v_i)}$ & Learnable vectors for node $v_i$ specified by the in-degree $deg^{-}(v_i)$\\
            $z^{+}_{deg^{+}(v_i)}$ & Learnable vectors for node $v_i$ specified by the in-degree $deg^{+}(v_i)$\\
            $Q_t$ & Query embeddings for globally temporal-spatial attention\\
            $K_t$ & Key embeddings for globally temporal-spatial attention\\
            $V_t$ & Value embeddings for globally temporal-spatial attention\\
            $W_q$ & Weight matrix for the query embeddings\\
            $W_k$ & Weight matrix for the key embeddings\\
            $W_v$ & Weight matrix for the value embeddings\\
            $E_e$ & Edge features for the edge $e$\\
            $W_e$ & Weight matrix for the edge features\\
            $c_{ij}$ & Shortest path features for the node pair $v_i$ and $v_j$\\
            $\mathbf{B}_{(i)}$ & Ouput features of $h$-th graph transformer layer\\
            $\mathbf{V}$ & Weight matrix for the multi-head graph transformer layer\\
            $\mathbf{H}_t$ & Final fused features at $t$-th time step\\
			\bottomrule[1.pt]
		\end{tabular}
	}
\end{table}

\section{Methodology}
\label{sec-method}
\subsection{Overview}
\begin{figure*}[h]
	\centering
    \includegraphics[width=0.9\textwidth,height=8cm]{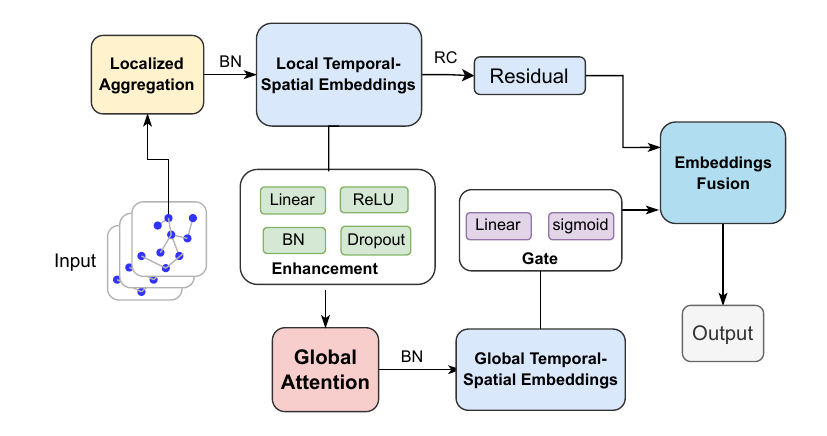}
	\caption{Overview of TSFusion.}
    \label{fig-overview}
\end{figure*}

In this work, we elaborate on the proposed architecture of the fusion model. As shown in Fig.~\ref{fig-overview}, 
the proposed model is composed of several vital components: temporal-spatial encoder (TSE), graph transformer encoder (GTransformer), and the adaptive fusion module (AFM). In the spatial-temporal encoder, it can jointly capture the temporal dynamics and local spatial dependencies of the traffic flow information encoded in each node feature vector. Alternatively, the graph transformer encoder can capture the global spatial dependencies between each node via the crafted attention mechanism. In the adaptive fusion module, the proposed model utilizes the gated mechanism to adaptively determine the relative importance between the TSE embeddings and GTransformer embeddings so as to let the model apply the message passing mechanism across multi-granularity levels, which is particularly suitable for modeling the complex traffic environment in the real world. 

\subsection{Locally Temporal-Spatial Aggregation}
In order to extract the localized structured information and temporal dynamics of traffic data across different time steps, as well as the geographical correlations among monitor stations, we deploy the temporal convolution with the graph-based message passing mechanism on the dynamic graphs. 

\begin{figure}[h]
	\centering
    \includegraphics[width=0.48\textwidth,height=5cm]{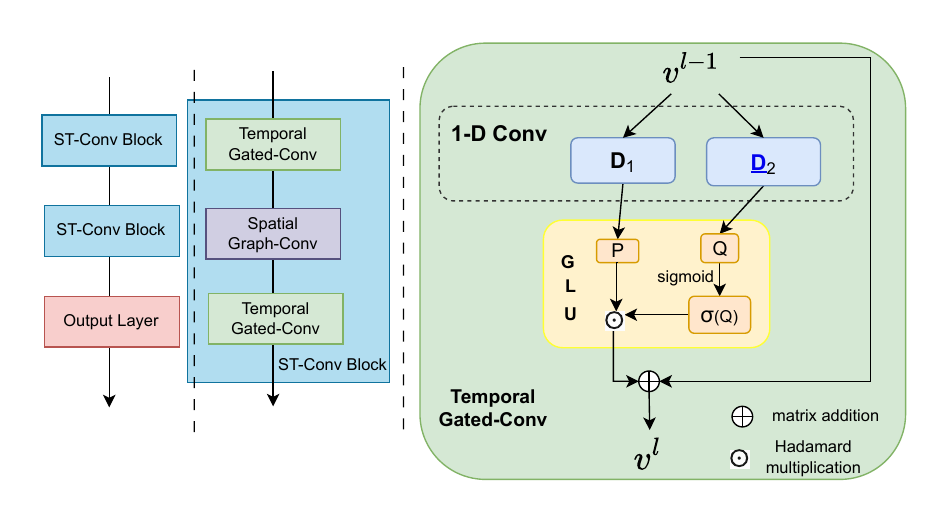}
	\caption{Details of the locally temporal-spatial aggregation.}
    \label{fig-stgcn}
\end{figure}

To capture the temporal traffic patterns, we utilize the 1-D causal convolutional layer followed by gated linear units (GLU) to produce the time-aware embeddings for each node. Specially, the 1-D causal convolution regards each node features as a length-$L$
sequence and use the convolutional filter $\mathcal{F}$ to convolve the node features at each time step only with its historical records and get its feature transformation $P$. We also deploy another 1-D causal convolutional layer on node features parallelly and get new features $Q$. After that, we utilize a GLU layer on the two features $P$ and $Q$, i.e., 
\begin{equation}
    C_{T}(\mathbf{X}_{t-\tau^{\prime}:t})=P_{t-\tau:t}\odot\sigma(Q_{t-\tau:t}).
\end{equation}
As for capturing the localized spatial patterns, we apply the graph convolutional layer on each frame of the dynamic graphs to aggregate traffic flow information according to the spatial relationship between each monitor station, i.e., 
\begin{equation}
    C_{S}(C_{T}(\mathbf{X}_{t}), \mathbf{A}_{t})=\sigma(\mathbf{\tilde{D}}_{t}^{-\frac{1}{2}}\tilde{\mathbf{A}}_{t}\tilde{\mathbf{D}}_{t}^{-\frac{1}{2}}C_{T}(\mathbf{X}_{t})\mathbf{W}^{(l+1)}),
\end{equation}
where $\mathbf{\tilde{D}}_{t}$ is the diagonal matrix, $\mathbf{\tilde{A}}_{t}=\mathbf{A}_t+\mathbf{I}$, $\sigma(\cdot)$ is the nonlinear activation function. It is worth noting that the input features $C_{T}(\mathbf{X}_{t})$ incorporate the temporal information in the dynamic traffic graphs. In the sequel, the proposed TSE is stacked by a ``sandwich" blocks structure in which the 1-D causal convolutional layer which is followed by a GLU layer is placed between two graph convolutional layer. After stacking several such blocks, we produce the time-aware node embeddings for each of the monitor stations, i.e., $(\mathbf{L}_{t-\tau},..., \mathbf{L}_{t})$. As a result, the locally spatial-temporal embeddings $(\mathbf{L}_{t-\tau},..., \mathbf{L}_{t})$ incorporate the temporal information via the temporal convolution and the locally structure information via the graph convolution. However, the bottleneck is that the graph convolutional layer only focus on preserving the similarities between the neighboring monitor stations and fails to capture the global structural patterns in the complex traffic graph data. To tackle this issue, we proposed to devise a special attention mechanism from the graph transformer to mine the temporal-spatial relationship between long-distance monitor stations in the dynamic traffic graph.

\subsection{Globally Temporal-Spatial Attention}
The GCN-based encoders can only capture the stationary spatial relationship between each monitor station of the traffic flow graph, which hinders the representation capability of the predictive model. Inspired by the graph transformer's powerful representative learning beyond the stationary topology, we craft a graph transformer block to utilize its special self-attention mechanism to efficiently mine long-range temporal-spatial relationships between each monitor station beyond the pre-defined geographical topology within traffic networks. 

In the graph transformer blocks, we first define the prior positional embeddings of the graph transformer blocks to inject the spatial information into the input embedding sequences. In this work, we introduce the extra graph-based feature engineering techniques, such as the node centralities into the node embeddings to produce the spatial-aware input features for the graph transformer blocks. As a result, the input feature sequences of the graph transformer blocks is:
\begin{equation}
    \tilde{\mathbf{L}}_{t,i}=\mathbf{L}_{t,i}+z^{-}_{deg^{-}(v_i)}+z^{+}_{deg^{+}(v_i)},
\end{equation}
where $z^+_i$ and $z^-_i$ are learnable vectors specified by the in-degree $deg^{-}(v_i)$ and out-degree $deg^{+}(v_i)$ respectively. Under these circumstances, we can incorporate the node importance signal into the self-attention mechanism of the graph transformer layer. 

\begin{figure}[h]
	\centering
    \includegraphics[width=0.45\textwidth,height=12cm]{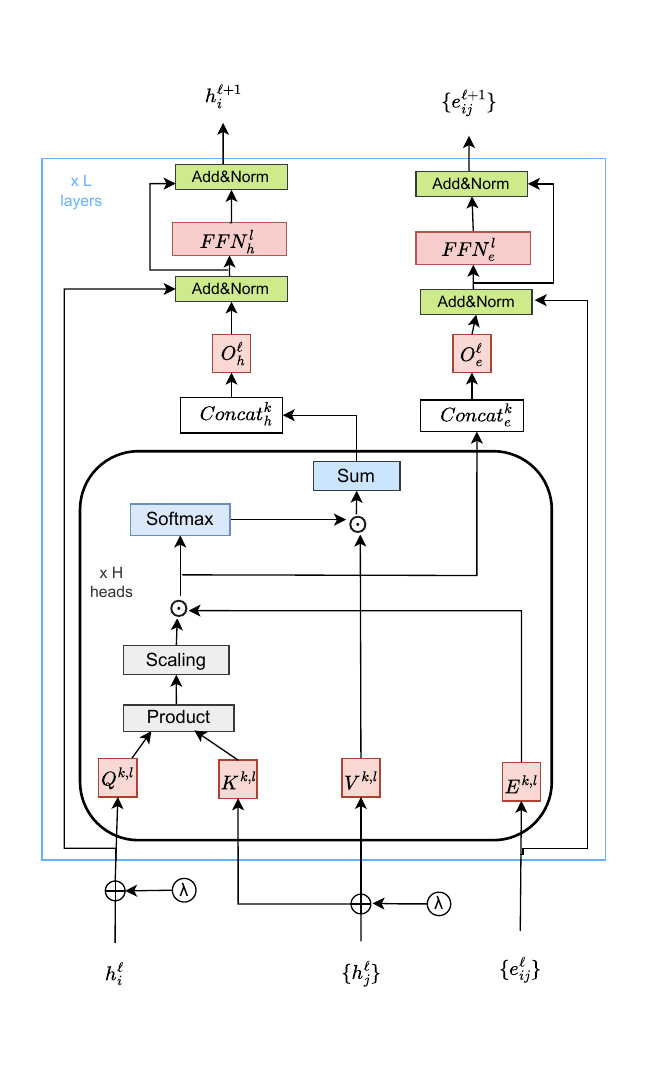}
	\caption{Details of the globally temporal-spatial attention.}
    \label{fig-graphtransformer}
\end{figure}

Next, we introduce the graph transformer layer to mine the global temporal-spatial signal of the dynamic traffic networks. The graph transformer layer will consider the potential connections between all the node pairs and utilize the special self-attention mechanism to mine the ``invisible" links in the traffic networks. Specially, we first adopt a feature enhancement layer $\textbf{FE}(\cdot)$ including the batch-normalization~\cite{BN} with dropout~\cite{Dropout} tricks to improve the representation capability of the embedding sequences $\tilde{\mathbf{L}}_t$ from the previous localized temporal-spatial module. After that, we project the localized embedding sequences in parallel into three high-dimensional spaces by three (multilayer perceptron) MLP layers, i.e., 
\begin{equation}
        \begin{split}
            Q_t=\tilde{\mathbf{L}}_t W_q,
            K_t=\tilde{\mathbf{L}}_t W_k,
            V_t=\tilde{\mathbf{L}}_t W_v,
        \end{split}
    \end{equation}
where $W_q$, $W_k$, $W_v$ are the weight matrices respectively. We also incorporate the edge features into the graph transformer layer by computing the average edge features between the shortest path $(v_i,e_1,e_2,...,e_r,v_j)$ of each node pair, i.e., 
\begin{equation}
    c_{ij}=\frac{1}{r}\sum_{e=1}^{r}E_e\cdot W_e^{\top},
\end{equation}
where $E_e$ is the edge features for the edge $e$, $W_e$ is the learnable weight matrix.
Then, the self-attention mechanism (the attention matrix $\mathbf{B}$) of the graph transformer layer is formulated as:
\begin{equation}
    \mathbf{B}_{ij}=\text{softmax}(\frac{Q_{t,i}\cdot K_{t,j}^{\top}+c_{ij}}{\sqrt{d}}),
\end{equation}
here, the softmax function is used to normalize the temporal-spatial dependencies, and the matrix $\mathbf{V}$ is the learnable weight matrix. To further improve the representation capability of the graph transformer layer, we adopt the multi-head attention (MHA) mechanism to jointly concatenate temporal-spatial signals from different perspectives, i.e.,
\begin{equation}
    \tilde{\mathbf{B}}=\text{Concat}[\mathbf{B}_{(1)}|\mathbf{B}_{(2)}|...| \mathbf{B}_{(h)}]\cdot\mathbf{V},
\end{equation}
where $\mathbf{B}_{(h)}$ is the $h$-th parallel graph transformer layer. It is worth noting that we adopt the layer normalization trick after each graph transformer layer to stabilize the training procedure.   

\subsection{Local and Global Features Fusion}
In this component, we further design a gated fusion mechanism in the adaptive fusion module to adaptively adjust the relative impacts from the locally and globally temporal-spatial patterns from temporal-spatial encoder and graph transformer encoders. In the adaptive fusion module, we choose the crafted \textit{gated residual connection} techniques to combine the transformed locally temporal-spatial patterns $\text{MLP}(\mathbf{L}_t)$ and globally temporal-spatial patterns $\text{MLP}(\tilde{\mathbf{B}})$, i.e.,    
\begin{equation}
    \begin{split}
        &Gate=\sigma(\text{MLP}(\tilde{\mathbf{B}}_t)\cdot W_{g}^{\top}+b_g), \\
        &\mathbf{H}_t=Gate\odot\text{MLP}(\tilde{\mathbf{B}}_t)+(1-Gate)\odot \text{MLP}(\mathbf{L}_t),
    \end{split}
\end{equation}
where $Gate$ measures the relative impacts of the global patterns $\text{MLP}(\tilde{\mathbf{B}})$ on the final fused features $\mathbf{H}$, $W_g$ and $b_g$ are the learnable parameters of the gated units, $\odot$ is the Hadamard product. Based on the local and global patterns fusion, our proposed model can adaptively encode the locally temporal-spatial patterns from the pre-defined geographical information of the traffic networks and ``invisible" globally temporal-spatial patterns observed from the graph transformer blocks into the final node embedding sequences. It is worth noting that we set the input features of the graph transformer encoder as the outputs from the previous temporal-spatial encoder instead of the raw nodal features due to the strong representation capability of the ``sandwich” blocks structure on the temporal-spatial space. After that, we deploy the mean square loss (MSE) to supervise the model training:
\begin{equation}
    \mathcal{L}=\sum_t \|\text{MLP}(\mathbf{H}_t)-\mathbf{X}_t\|^2.
\end{equation}

\section{Experiments}
In this section, we evaluate the performance of our proposed method on two real-world datasets and aim to answer the following questions:
\begin{itemize}
    \item \textbf{RQ1}: How is the predictive performance of our proposed method compared to various strong baselines on the graph-based traffic flow prediction task?
    \item \textbf{RQ2}: How are the crafted modules' impact on the model performances? 
    \item \textbf{RQ3}: How does the proposed method perform with regard to different time periods?
    \item \textbf{RQ4}: How does the proposed method perform under different noise levels?
\end{itemize}

\subsection{Dataset Descriptions}
We deploy the proposed model and all the baselines on two real-world dynamic traffic graph datasets, i.e., PeMSD4 and PeMSD8 from California, where they are collected from the California Department of Transportation in real time every 30 seconds. Each dataset contains a series of vital features of traffic flow observations and geographical information with different time steps. All the traffic flow observations are gathered every 5 minutes. The Caltrans Performance Measurement System (PeMS) has more than 39,000 detectors deployed on the highway in the major metropolitan areas in California. The topology information, which incorporates the relative locations among the monitor stations are recorded in the two datasets. In the experiments, we consider three typical traffic flow measurements, including total flow, average speed, and average occupancy. Next, we describe in more detail of these two datasets. 

PeMSD4 contains the traffic flow information in the San Francisco Bay Area, containing 3848 monitor stations on 29 roads. The period of this dataset is from January to February in 2018. We choose data on the first 50 days as the training set, and the remaining as the testing set. PeMSD8 contains the traffic flow information in San Bernardino from July to August in 2016, which contains 1979 monitor stations on 8 roads. The data on the first 50 days is used as the training set, and the data on the last 12 days refers to the testing set.  

During the data pre-processing phase, the topology information of the two real-world dynamic traffic graphs are formulated as a dynamic weighted directed graph computed based on Eqn.~\ref{eqn-data-preprocessing}. We summarize the descriptive statistics of the traffic graph datasets in Tab.~\ref{tab-dataset}.

\begin{table}[h]
	\centering
	\caption{Dataset statistics.}
	\label{tab-dataset}
	\resizebox{1.\columnwidth}{!}{%
		\begin{tabular}{c|cc}
			\toprule[1.pt]
			Datasets & PeMSD4 & PeMSD8 \\
			\hline
			Time duration      &$1$ month  &$1$ month   \\
            Time step          &$5$ minutes&$5$ minutes \\
			Sample size        &$16992$ & $17856$\\
            \#Roads            &$29$    &$8$\\
            \#Monitor stations &$307$ &$170$ \\
            Graph type         &Undirected Weighted graph &Undirected Weighted graph \\
            \# Node features   &$3$ &$3$ \\
			\bottomrule[1.pt]
		\end{tabular}
	}
\end{table}

\subsection{Experimental Settings}
All the experiments are compiled and tested on a Windows 10 desktop (CPU: Intel(R) Core i7, GPU: NVIDIA GeForce RTX 4090), 32 GB RAM. The implementation is based on Python 3.8, and PyTorch 1.10.0. In order to eliminate the atypical traffic phenomenon, we only adopted the workday traffic data in all the experiments. We removed some redundant monitor stations to ensure the distance between any adjacent station is longer than 3.5 miles. Hence, there are $307$ stations in the PeMSD4 and $170$ stations in the PeMSD8. Moreover, the traffic data are aggregated every $5$ minutes, so each monitor station contains $288$ data points per day. The missing values are filled by the linear interpolation trick. In addition, the data are transformed by zero-mean normalization, i.e.,  $x^\prime=x-\bar{x}$ where $\bar{x}$ is the mean value of $x$. We executed a grid search strategy to search for the best parameters on the validation set. All the experiments used $60$ minutes as the historical time window and $12$ observed data points for traffic forecasting conditions in the next $15$, $30$, and $45$ minutes. 

\subsection{Baseline \& Metrics}
We compare our proposed method with seven baselines, including statistical methods, machine learning based methods, and deep learning based methods. The detailed descriptions are as follows:

\begin{figure*}[h]
	\centering
	\begin{subfigure}[b]{0.49\textwidth}
		\centering
		\includegraphics[width=\textwidth,height=6.5cm]{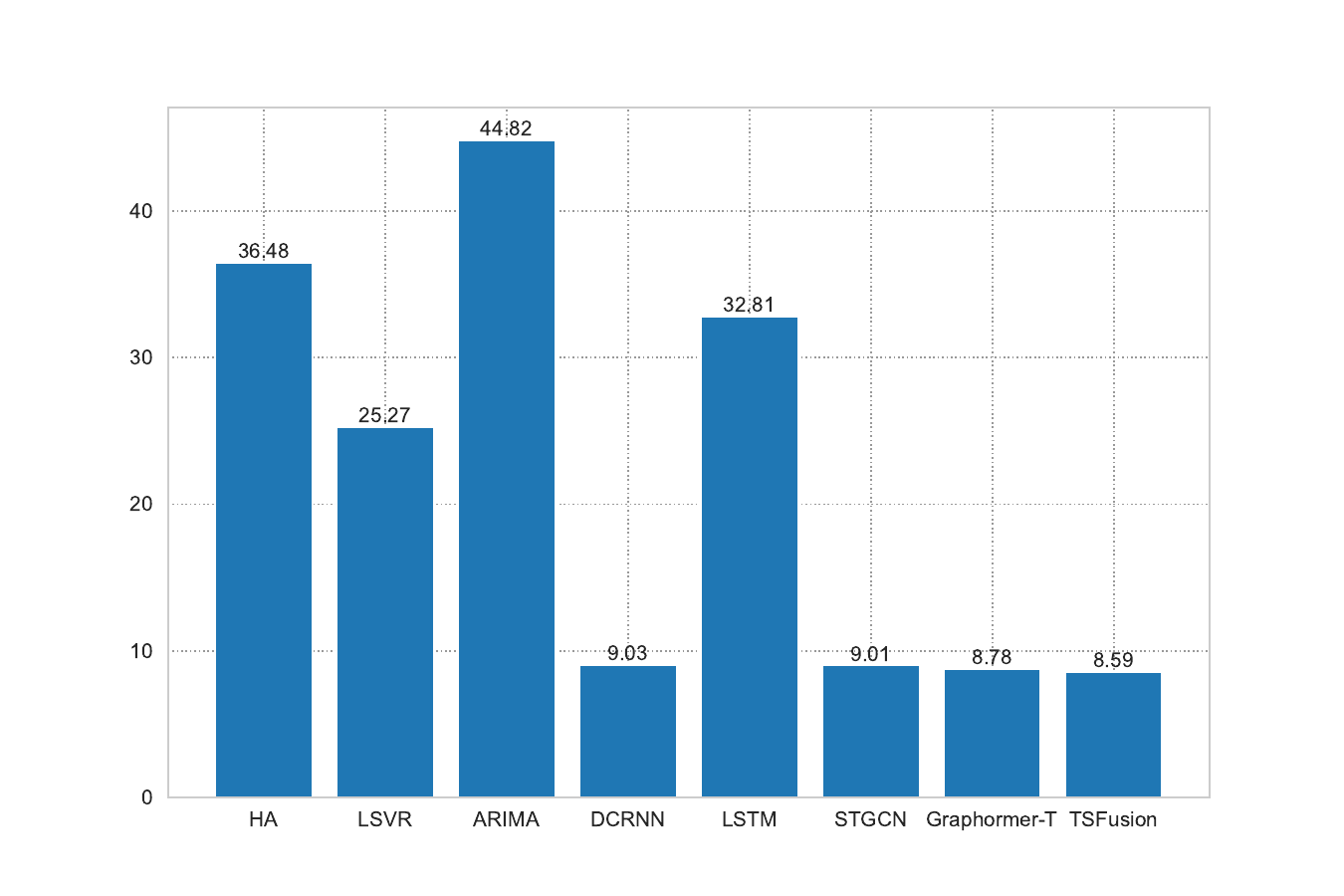}
		\caption{MAE for PeMSD4}
        \label{fig-P4-MAE}
	\end{subfigure}
	\hfill
	\begin{subfigure}[b]{0.49\textwidth}
		\centering
		\includegraphics[width=\textwidth,height=6.5cm]{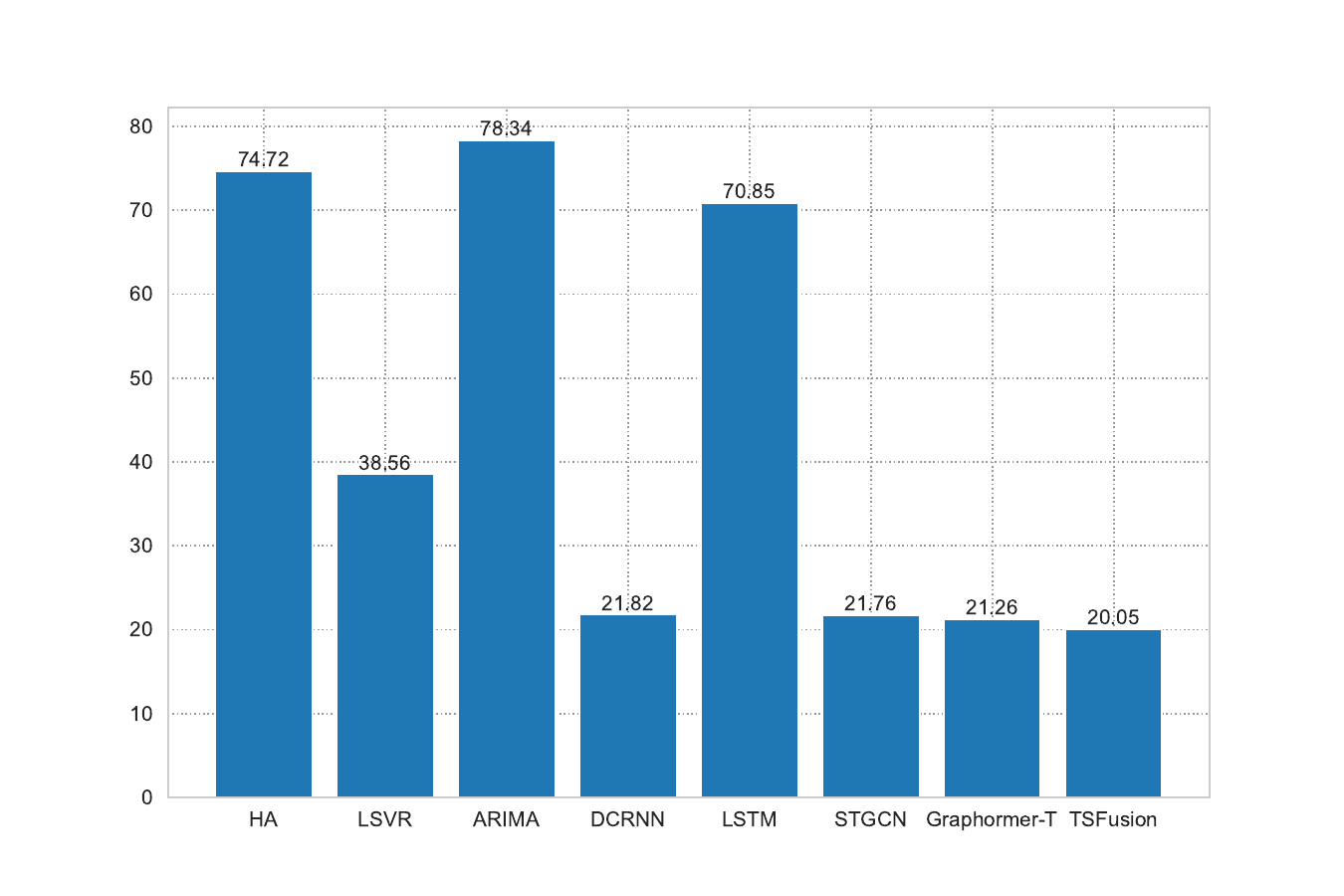}
		\caption{RMSE for PeMSD4}
        \label{fig-P4-RMSE}
	\end{subfigure}
    \hfill
	\begin{subfigure}[b]{0.49\textwidth}
		\centering
		\includegraphics[width=\textwidth,height=6.5cm]{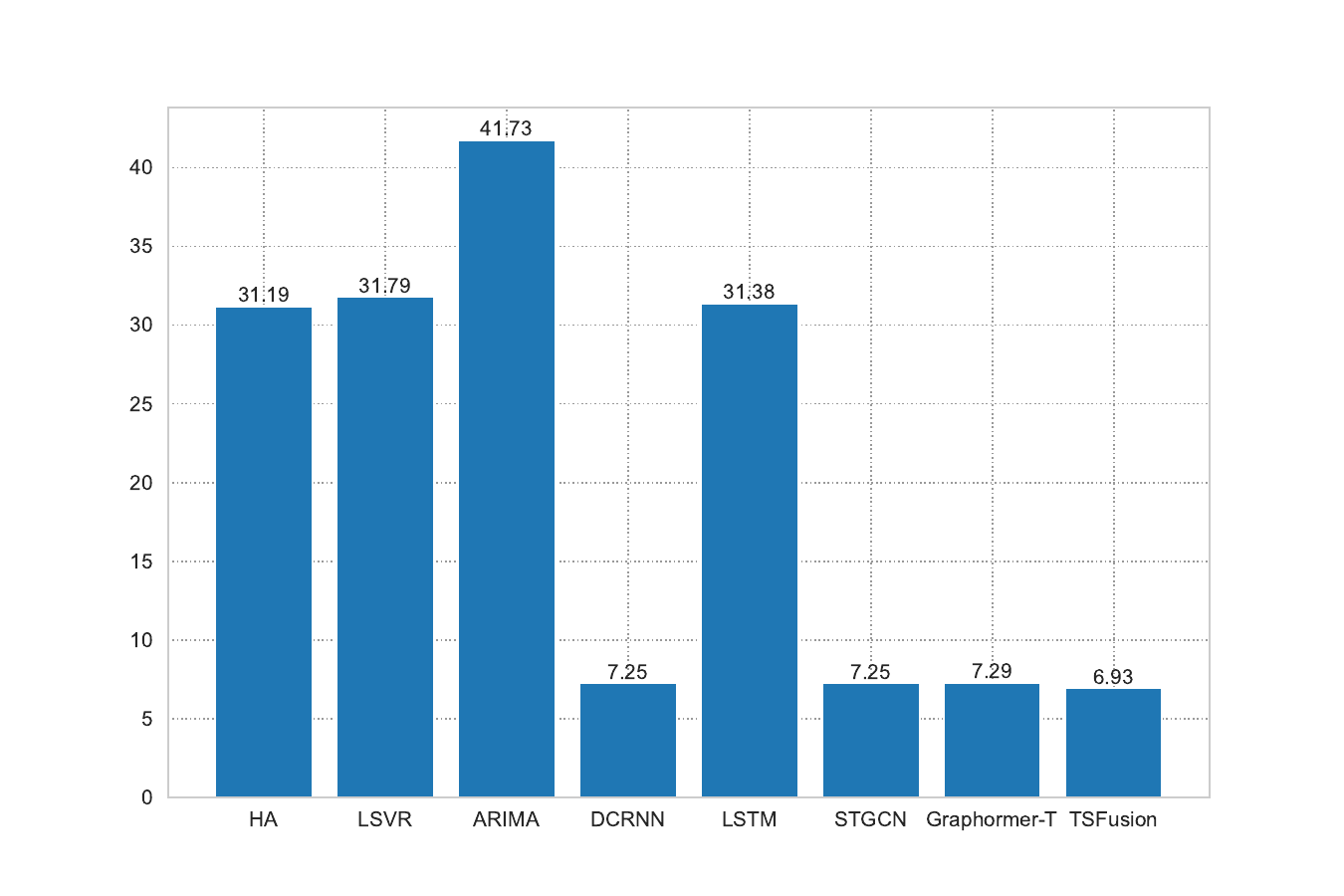}
		\caption{MAE for PeMSD8}
        \label{fig-P8-MAE}
	\end{subfigure}
    \hfill
	\begin{subfigure}[b]{0.49\textwidth}
		\centering
		\includegraphics[width=\textwidth,height=6.5cm]{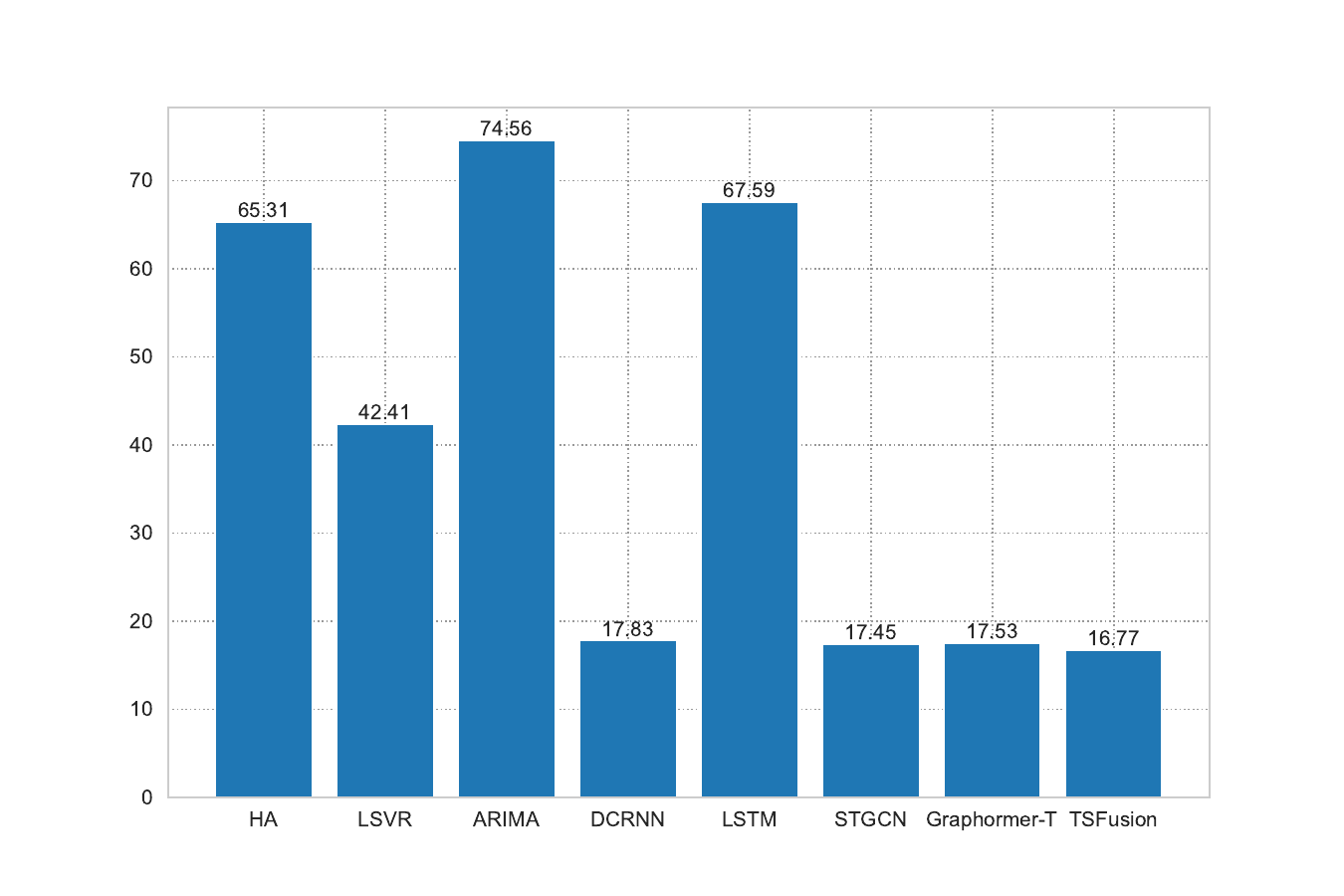}
		\caption{RMSE for PeMSD8}
        \label{fig-P8-RMSE}
	\end{subfigure}
	\caption{Model comparison results on two traffic datasets.}
	\label{fig-exp1}
\end{figure*}

\begin{itemize}
    \item \textbf{HA}: (Historical Average method), the forecasting result is the average value of the same time in the training data.
    \item \textbf{ARIMA}: (Auto Regressive Integrated Moving Average) is a commonly used model for time series prediction, which can effectively extract the long-term dependency of time series.
    \item \textbf{LSVR}: (Linearized Support Vector Regression) is a machine learning framework that uses a Support Vector Machine for regression problems.
    \item \textbf{LSTM}: is a specialized recurrent neural network (RNN) variant that uses gated memory cells to solve long-term dependency problems by selectively retaining or discarding information over time sequences.
    \item \textbf{DCRNN}: deep learning framework for traffic forecasting that captures the spatial dependency using bidirectional random walks on the graph, and the temporal dependency using the encoder-decoder architecture with scheduled sampling.
    \item \textbf{STGCN}:  is a spectral-based graph convolution model with gated linear temporal convolution to capture spatial-temporal dependencies.
    \item \textbf{Graphormer-T}: is a transformer-based deep learning framework that effectively integrates topology information through centrality, spatial, and edge encoding techniques with a specialized self-attention mechanism. \textit{Since the original Graphormer does not contain the temporal information, during the data preprocessing phase, we choose to concatenate the nodal features from previous $k$ time steps as the input for model training. We then rename this variant as} \textbf{Graphormer-T}.  
\end{itemize}
We use the two typical metrics, i.e., Mean Absolute Error (MAE) and Rooted Mean Squared Error (RMSE) to measure the traffic forecasting performances, which are defined as:
\begin{equation}
    \text{MAE}=\frac{\sum_{i=1}^{N}|\mathbf{X}_i-\hat{\mathbf{X}_i}|}{N},
\end{equation}
\begin{equation}
    \text{RMSE}=\sqrt{\frac{\sum_{i=1}^{N}(\mathbf{X}_i-\hat{\mathbf{X}_i})^2}{N}},
\end{equation}
where $\hat{\mathbf{X}}_i$ is the predicted results for the traffic flow information of the $i$-th monitor station, $\mathbf{X}_i$ is the ground truth. 

\subsection{Traffic Flow Forecasting Performances}
In this section, we evaluate our proposed methods with other baselines on two typical traffic graph datasets (PeMSD4 and PeMSD8). We evaluated each model $10$ times and took the average values as the final predicted performances, which are shown in Fig.~\ref{fig-P4-MAE}, Fig.~\ref{fig-P4-RMSE}, Fig.~\ref{fig-P8-MAE}, and Fig.~\ref{fig-P8-RMSE}. We use MAE and RMSE metrics to evaluate the traffic flow forecasting performance of all the predictive methods. We first compare our proposed method with other baselines on the PeMSD4 dataset presented in Fig.~\ref{fig-P4-MAE} and \ref{fig-P4-RMSE}. The results demonstrate that our proposed method outperforms all the baselines in the metrics of MAE and RMSE. 

\begin{table*}[h]
	\centering
	\caption{Forecasting performances on varying time durations.}
	\label{tab-exp-varying-time}
	\resizebox{1.8\columnwidth}{!}{%
		\begin{tabular}{c|cc|cc}
			\toprule[1.pt]
			\multirow{2}*{Model} & \multicolumn{2}{c}{PeMSD4 ($15$min/$30$min/$45$min)} & \multicolumn{2}{|c}{PeMSD8 ($15$min/$30$min/$45$min)} \\
			
			      &MAE  &RMSE & MAE & RMSE   \\
            \hline 
            HA & $36.48$/$36.48$/$36.48$ & $74.72$/$74.72$/$74.72$ & $31.19$/$31.19$/$31.19$ & $65.31$/$65.31$/$65.31$\\
            LSVR & $25.27$/$27.65$/$29.87$ & $38.56$/$40.38$/$42.36$ & $31.79$/$32.98$/$33.04$ & $42.41$/$43.56$/$44.38$\\
            ARIMA & $44.82$/$45.16$/$47.69$ & $78.34$/$79.54$/$82.37$ & $41.73$/$43.55$/$44.68$ & $74.56$/$75.87$/$78.68$\\
            DCRGNN & $9.03$/$11.14$/$13.34$ & $21.82$/$26.41$/$31.29$ & $7.25$/$9.02$/$10.88$ & $17.83$/$21.94$/$26.03$\\
            LSTM & $32.81$/$33.25$/$33.82$ & $70.85$/$71.81$/$73.01$ & $31.38$/$32.83$/$33.69$ & $67.59$/$68.41$/$69.44$\\
            STGCN & $9.01$/$10.86$/$12.87$ & $21.76$/$25.81$/$30.30$ & $7.25$/$8.82$/$10.54$ & $17.45$/$21.26$/$25.17$\\
            Graphormer-T & $8.78$/$10.63$/$12.70$ & $21.26$/$25.99$/$30.53$ & $7.29$/$8.89$/$10.55$ & $17.53$/$21.24$/$25.17$\\
            TSFusion & $\mathbf{8.59/9.36/11.23}$ & $\mathbf{20.05/23.32/28.36}$ & $\mathbf{6.93/8.61/10.42}$ & $\mathbf{16.77/20.81/24.92}$\\
			\bottomrule[1.pt]
		\end{tabular}
	}
\end{table*}

Compared with the machine learning and deep learning based methods, the traditional statistical methods, such as HA and ARIMA, because they highly rely on strong assumptions on the linearized feature space and are prone to be underfitting on the training data. On the other hand, the machine learning based methods, i.e., LSVR, can only achieve limited incremental performances due to their strong assumptions on the feature space and shallow architectures, which will hinder their representation capability. As a consequence, the statistical approaches together with machine learning based methods have a fatal representation learning bottleneck and are unable to capture the complex nonlinear temporal-spatial traffic flow patterns in the traffic graph datasets. 

On the other hand, the graph convolution based methods, such as DCRNN, STGCN achieve better performances than other plain time series models such as HA, ARIMA and LSTM demonstrate that the graph convolution can effectively capture the topological structural information of the traffic graph data. Moreover, the recurrent neural units of DCRNN and the 1D causal convolutional layer of STGCN can simultaneously model the nonlinear relations between the traffic flow patterns at different time steps, thus leads to strong temporal-spatial representation capability. We also include a transformer-based model, i.e., Graphormer~\cite{graphormer}, and observe that our proposed model slightly outperforms the transformer-only model, which indicates that the transformer-only models are prone to overfit on the global temporal-spatial patterns and thus lead to suboptimal results. This phenomenon happened possibly due to the self-attention mechanism of the transformer layer, which mines too much ``unseen" topological information and severely distorts the original geographical relations between the monitor stations. 

Next, by comparing our proposed method with other baselines on PeMSD8, it is observed that our proposed method can consistently achieve the best performance among all the above-mentioned methods and datasets, which demonstrates the validity of our insight, i.e., the real-world traffic flow patterns are encoded into a highly complex feature space and single-grained encoder is insufficient to learn a high-quality representation learning for such complicated scenarios. Thus, we need to fuse multi-grained temporal-spatial correlations simultaneously for better traffic flow forecasting performance.   

\subsection{Forecasting Performances on Varying Durations}
In this section, we further compare our fusion-based method with other baselines on the two traffic graph datasets for varying time durations of the next traffic flow generation in Tab.~\ref{tab-exp-varying-time}. Without loss of generality, we still adopt the MAE and RMSE to measure the forecasting performance of all the methods on varying time intervals. It is observed that the prediction errors of all models increase as the time duration increases, which is in line with the general rule of time series prediction: the further the prediction time is, the higher the uncertainty is. On the other hand, our proposed model still consistently demonstrates superior performance across various time durations, and this advantage is particularly pronounced in long-term predictions (up to 45 minutes). Overall, based on the forecasting results from Fig.~\ref{fig-exp1} and Tab.~\ref{tab-exp-varying-time}, we summarize that by integrating the multi-grained temporal-spatial patterns of the traffic graph datasets, our proposed method can achieve consistently better representation learning than the single-grained encoders such as the convolution-based methods and transformer-based methods.  

\begin{table}[htp]
	\centering
	\caption{Ablation studies.}
	\label{tab-exp-ablation}
	\resizebox{0.8\columnwidth}{!}{%
		\begin{tabular}{c|cc}
			\toprule[1.pt]
			\multirow{2}*{Model} & \multicolumn{2}{c}{PeMSD4 ($15$min)} \\
			
			      &MAE  &RMSE   \\
            \hline 
            TSFusion-nG    & $9.02$ & $22.36$\\
            TSFusion-nL    & $8.79$ & $22.10$\\
            TSFusion-nFE   & $9.01$ & $21.21$\\
            TSFusion-nGate & $8.75$ & $20.98$\\
            TSFusion-nRes  & $8.55$ & $20.35$\\
            TSFusion       & $8.35$ & $19.82$\\
			\bottomrule[1.pt]
		\end{tabular}
	}
\end{table}

\subsection{Ablation Study}
In this section, we present our ablation study experiments of our proposed method to validate the effectiveness of each component mentioned in Tab.~\ref{tab-exp-ablation}. Specifically, there are five variants of our proposed method:
\begin{itemize}
    \item \textbf{TSFusion-nG}: is the TSFusion without the globally temporal-spatial attention mechanism.
    \item \textbf{TSFusion-nL}: is the TSFusion without the locally temporal-spatial aggregation mechanism.
    \item \textbf{TSFusion-nFE}: is the TSFusion without the feature enhancement layer.
    \item \textbf{TSFusion-nGate}: is the TSFusion without the gated fusion mechanism.
    \item \textbf{TSFusion-nRes}: is the TSFusion without the skip-connection layer. 
\end{itemize}

\begin{figure*}[!th]
	\centering
	\begin{subfigure}[b]{0.325\textwidth}
		\centering
		\includegraphics[width=\textwidth,height=4.5cm]{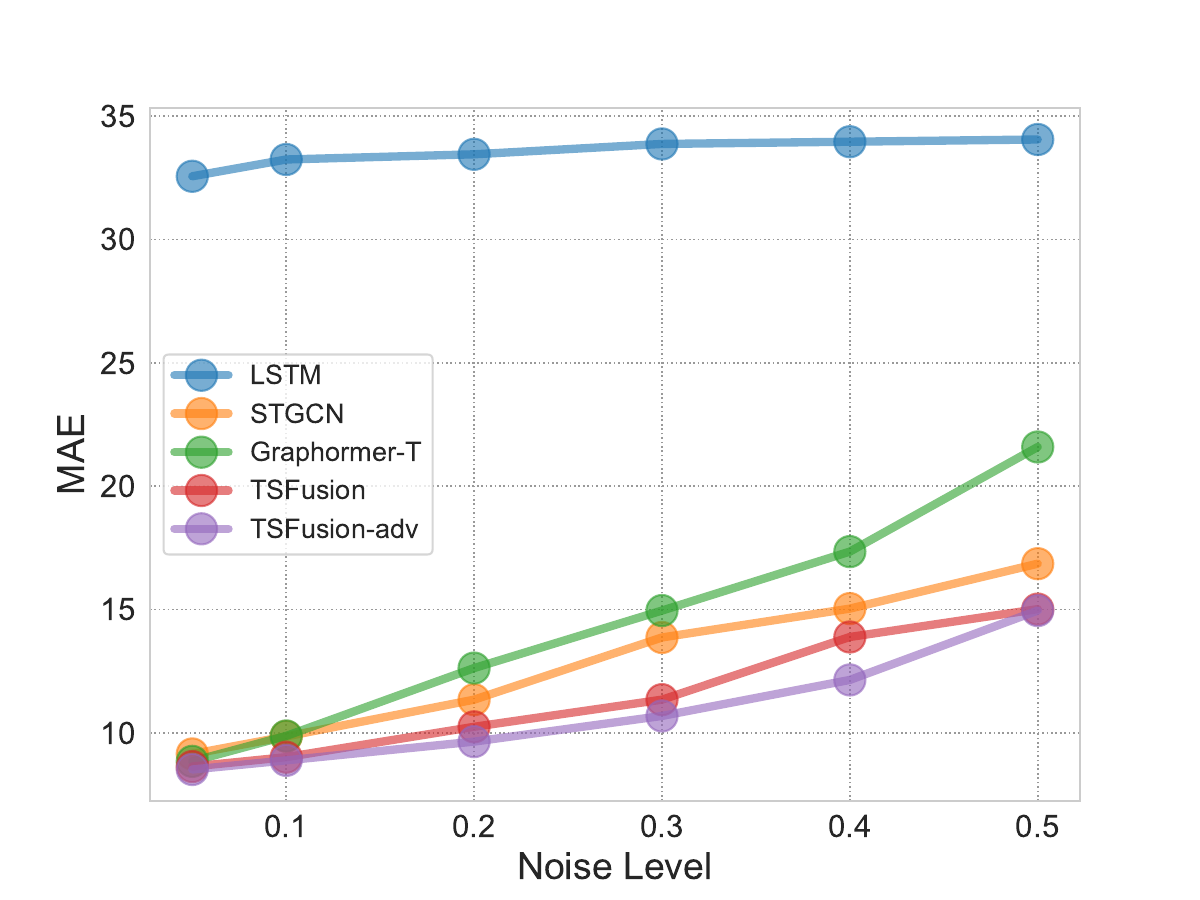}
		\caption{Gaussian noise}
        \label{fig-gaussian}
	\end{subfigure}
	\hfill
	\begin{subfigure}[b]{0.325\textwidth}
		\centering
		\includegraphics[width=\textwidth,height=4.5cm]{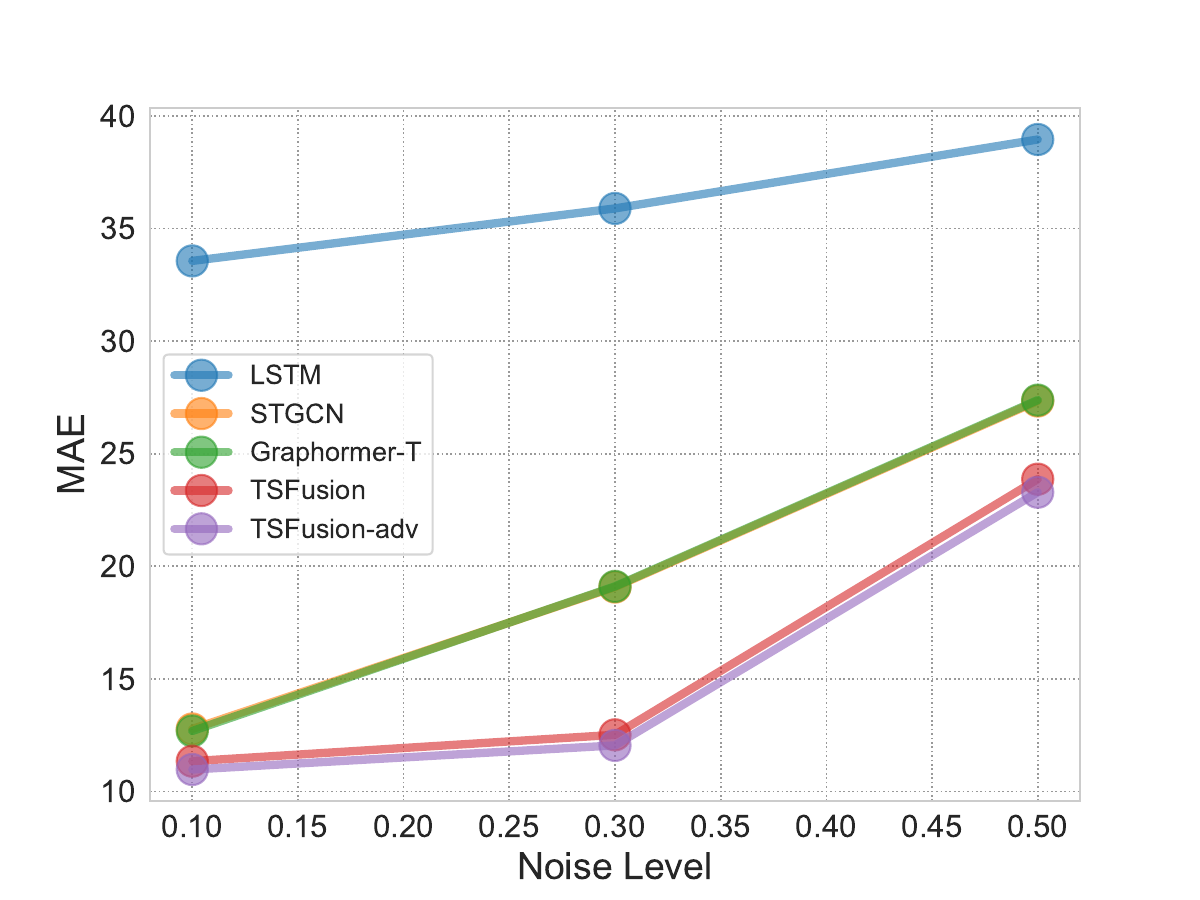}
		\caption{Missing data}
        \label{fig-missing}
	\end{subfigure}
    \hfill
	\begin{subfigure}[b]{0.325\textwidth}
		\centering
		\includegraphics[width=\textwidth,height=4.5cm]{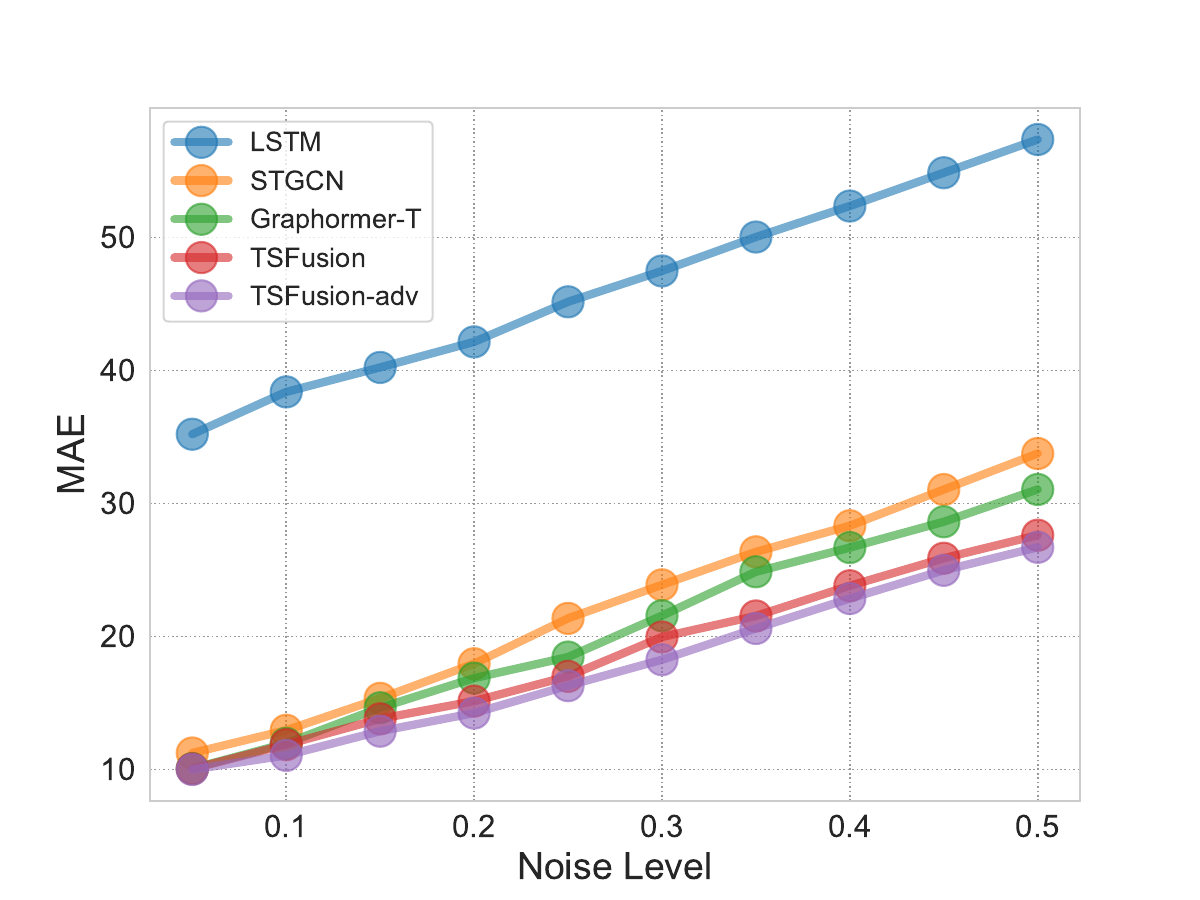}
		\caption{Adversarial samples}
        \label{fig-adv}
	\end{subfigure}
	\caption{Model comparison results on the two traffic datasets.}
	\label{fig-exp-robust}
\end{figure*}

As shown in Tab.~\ref{tab-exp-ablation}, it is observed that TSFusion without the globally temporal-spatial attention mechanism performs worse than \textbf{TSFusion-nL}, which indicates that the potential global temporal-spatial patterns are more important than the observed local temporal-spatial patterns for precise traffic flow predictions. This phenomenon is also coincide with our preliminary findings that unconnected nodes share the same function (for example, two nodes share the same traffic flow pattern if they both belong to the ``transportation hub"). On the other hand, by comparing \textbf{TSFusion-nFE} and \textbf{TSFusion}, we draw the conclusion that the feature enhancement layer is a vital component of our proposed model due to the suboptimal representation capability of the raw output features from the temporal-spatial blocks. Moreover, by comparing \textbf{TSFusion-nGate} with the full version, it is demonstrated that the gated fusion mechanism can rationally determine the relative importance between the local embeddings $\mathbf{L}_t$ and global embeddings $\tilde{\mathbf{B}}_t$ and thus boost the forecasting accuracies of the traffic flow prediction system. Lastly, it is observed that our full version slightly outperforms \textbf{TSFusion-nRes}, indicating that the introduced residual blocks are less important than the previously mentioned designs. It makes sense since the other four designs directly influence the quality of the mined local and global temporal-spatial patterns with their coordination in the final representation learning.

\subsection{Robustness Analysis}
In this section, we analyze the robustness of our proposed model with other typical baselines against three typical noises in the traffic graph datasets, i.e., Gaussian noise, missing data and adversarial attacks. 

\subsubsection{Gaussian Noise}
Under this scenario, we randomly inject the Gaussian noise features $\epsilon_t$ into the nodal attributed matrices $\mathbf{X}_t$ with different noise levels to mimic the possible data noises collected from the complex environment. The attributed matrix with Gaussian noise can be formulated as:
\begin{equation}
    \mathbf{X}^{g}_{t}=\mathbf{X}_t+\eta\odot\epsilon_t,
\end{equation}
where $\eta\in\{0.1,0.2,0.3,0.4,0.5\}$ measures the noise degree. We present the robust performances of our proposed method with four typical baselines under different noise levels in Fig.~\ref{fig-gaussian}. It is worth noting that \textbf{TSFusion-adv} is \textbf{TSFusion} augmented with the adversarial training~\cite{AdversarialTraining} on the feature space to further boost the adversarial robustness of the proposed method. The experiment results in Fig.~\ref{fig-gaussian} depict that our proposed method can achieve the best robust performance compared to the other three baselines when the node attribute matrices are polluted by random Gaussian noises. On the other hand, if we introduce the adversarial training technique when training \textbf{TSFusion}, we can slightly improve the robustness of \textbf{TSFusion} against the Gaussian noise.   

\subsubsection{Missing Data}
For this case, we randomly remove $10\%$, $30\%$, $50\%$ of the node features and set the missing data to zero during the data preprocessing phase to mimic the missing data problem in the real world, and the experiment results are presented in Fig.~\ref{fig-missing}. We observe that there exists a significant performance gap between our proposed method with other baselines, which indicates that our multi-grained graph representation learning can strongly defend against the missing data problem in the node feature space. Moreover, the phenomenon that the performances of \textbf{TSFusion} and \textbf{TSFusion-adv} are too close demonstrates that there is a lack of significant evidence that adversarial training tricks can handle the missing data problem well for traffic flow prediction.   

\subsubsection{Adversarial Attacks}
Lastly, we investigate the adversarial robustness of our proposed method against adversarial attacks on the feature space of the traffic graph dataset and present the experiment results in Fig.~\ref{fig-adv}. The results indicate that considering the multi-grained temporal-spatial traffic flow patterns can improve the adversarial robustness of the forecasting system to some extent, which paves the way for future research on the security of the deep learning based traffic flow prediction.

\section{Conclusion \& Future Work}
In this work, we propose a multi-grained temporal-spatial traffic flow prediction method that adaptively integrates the locally temporal-spatial graph convolutional encoders with globally temporal-spatial graph enhanced transformer encoders to enhance the representation capability and robustness. More specifically, in the locally temporal-spatial aggregation module, we adopt the causal convolution to produce the time-aware features and utilize the graph convolutional layers to capture the localized geographical relationship between each monitor station. On the other hand, to mine the potential connection between the unconnected traffic nodes which share a similar function (transportation hub, business center or residence zone), we adopt the self-attention mechanism of the graph enhanced transformer layer to globally search for the ``unseen" connections for each traffic node pair. To scientifically fuse the temporal-spatial patterns from different views, we adopt the gated fusion units to automatically determine the relative importance between local and global views for high-quality and robust representation learning. Extensive experiments demonstrate that our proposed fusion method can achieve the best forecasting accuracies among the strong baselines and can defend against the random noises, missing data, and adversarial attacks problem in the feature space of the traffic graph dataset.

In the future, we will explore the advanced pre-trained graph-based foundation models~\cite{liu2025graph} to further boost the forecasting accuracy of the current traffic flow predicting system. Moreover, we will investigate the adversarial robustness of the current traffic flow predicting system in depth and refer to the robust structural learning techniques to arm the proposed forecasting method with more advanced defense mechanisms.

\bibliographystyle{IEEEtran}
\bibliography{citation}

\end{document}